\documentclass{bmvc2k}
\usepackage{ulem, multirow, amssymb,tabularx,paralist,array,tabu}

\title{Gated Fusion Network for Joint Image Deblurring and Super-Resolution}

\addauthor{Xinyi Zhang}{http://xinyizhang.tech}{1}
\addauthor{Hang Dong}{dhunter@stu.xjtu.edu.cn}{1}
\addauthor{Zhe Hu}{zhe.hu@hikvision.com}{2}
\addauthor{Wei-Sheng Lai}{wlai24@ucmerced.edu}{3}
\addauthor{Fei Wang}{wfx@mail.xjtu.edu.cn}{1}
\addauthor{Ming-Hsuan Yang}{mhyang@ucmerced.edu}{34}

\addinstitution{
National Engineering Laboratory for \\
Vision Information Processing and Application, Xi\rq an Jiaotong University, \\ 
Xi\rq an, Shaanxi 710049, China
}
\addinstitution{
Hikvision Research
}
\addinstitution{
University of California, Merced
}
\addinstitution{
Google Cloud
}

\runninghead{Zhang, Dong, Hu, Lai, Wang, Yang}{GFN for Deblurring and SR}

\newcommand{\figref}[1]{Figure~\ref{fig:#1}}
\newcommand{\tabref}[1]{Table~\ref{tab:#1}}

\makeatletter
\newcommand{\thickhline}{%
    \noalign {\ifnum 0=`}\fi \hrule height 1pt
    \futurelet \reserved@a \@xhline
}
\newcolumntype{"}{@{\hskip\tabcolsep\vrule width 1pt\hskip\tabcolsep}}
\makeatother

\begin{document}

\maketitle

\begin{abstract}
Single-image super-resolution is a fundamental task for vision applications to enhance the image quality with respect to spatial resolution. 
If the input image contains degraded pixels, the artifacts caused by the degradation could be amplified by super-resolution methods.
Image blur is a common degradation source.
Images captured by moving or still cameras are inevitably affected by motion blur due to relative movements between sensors and objects. 
In this work, we focus on the super-resolution task with the presence of motion blur.
We propose a deep gated fusion convolution neural network to generate a clear high-resolution frame from a single natural image with severe blur. 
By decomposing the feature extraction step into two task-independent streams, the dual-branch design can facilitate the training process by avoiding learning the mixed degradation all-in-one and thus enhance the final high-resolution prediction results.
Extensive experiments demonstrate that our method generates sharper super-resolved images from low-resolution inputs with high computational efficiency.
\end{abstract}

\section{Introduction}
\label{sec:intro}
\begin{figure}[tb]
\centering
\begin{minipage}[b]{0.477\textwidth}
\subfigure[Blurry low-resolution input]{
\includegraphics[width=1\textwidth]{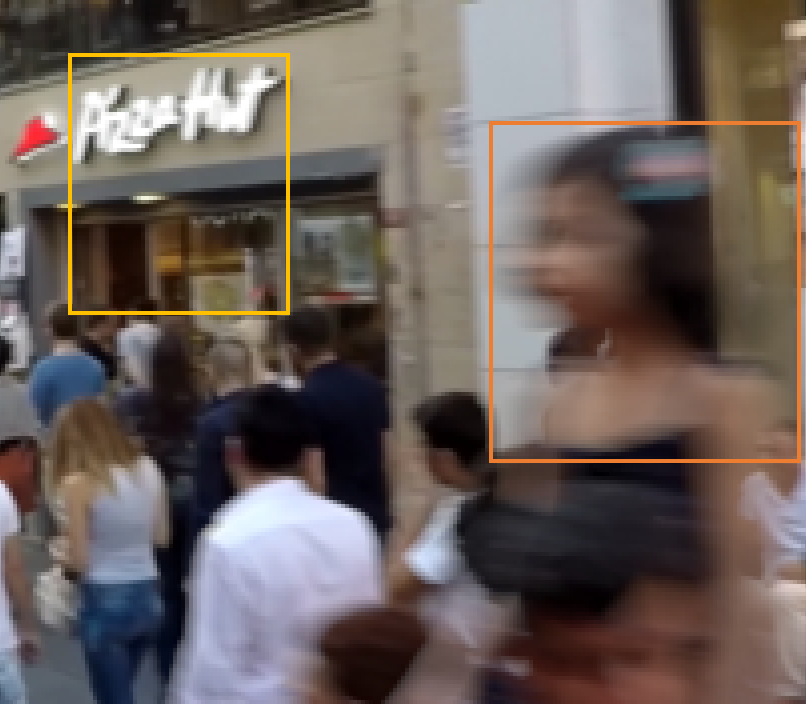} 
\label{fig:1_all}
}
\end{minipage}
%
\begin{minipage}[b]{0.513\textwidth}
\begin{minipage}[b]{0.32\textwidth}
\subfigure[Input patch]{
\includegraphics[width=1\textwidth]{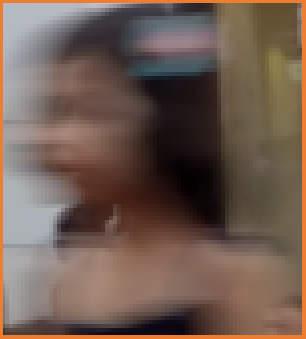} 
\label{fig:1_sr_input}
}
\end{minipage}
\begin{minipage}[b]{0.32\textwidth}
\subfigure[EDSR~\cite{EDSR}]{
\includegraphics[width=1\textwidth]{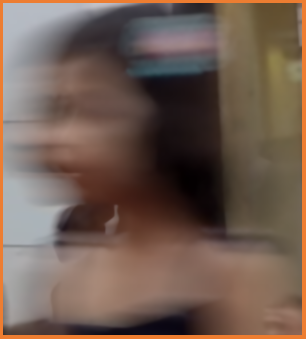} 
\label{fig:1_sr}
}
\end{minipage}
\begin{minipage}[b]{0.32\textwidth}
\subfigure[Ours]{
\includegraphics[width=1\textwidth]{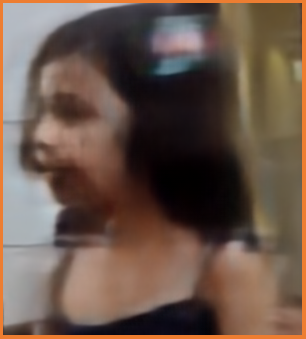}
\label{fig:1_sr_ours}
}
\end{minipage}
\vfill
\begin{minipage}[b]{0.32\textwidth}
\subfigure[Input patch]{
\includegraphics[width=1\textwidth]{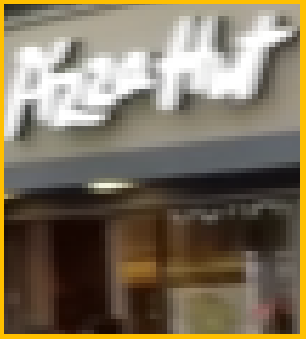} 
\label{fig:1_db_input}
}
\end{minipage}
\begin{minipage}[b]{0.32\textwidth}
\subfigure[DeepDeblur~\cite{deepdeblur}]{
\includegraphics[width=1\textwidth]{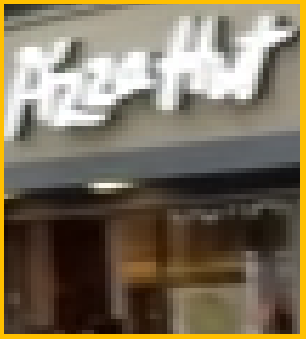} 
\label{fig:1_db}
}
\end{minipage}
\begin{minipage}[b]{0.32\textwidth}
\subfigure[Ours]{
\includegraphics[width=1\textwidth]{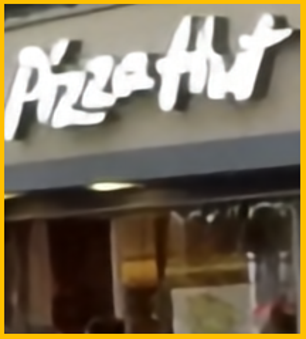} 
\label{fig:1_db_ours}
}
\end{minipage}
\end{minipage}
\caption{
\textbf{Examples of joint image deblurring and super-resolution.}
While the state-of-the-art SR algorithm~\cite{EDSR} cannot reduce the non-uniform blur in the input image due to the simple assumption of the bicubic downsampling, the top-performing non-uniform deblurring algorithm~\cite{deepdeblur} generates clear results but with few details.
In contrast, the proposed method generates a clear and high-resolution image with more details.
}
\label{fig:1}
\end{figure}

Single-image super-resolution (SISR) aims to restore a high-resolution (HR) image from one low-resolution (LR) frame, such as those captured from surveillance and mobile cameras.
The generated HR images can improve the performance of the numerous high-level vision tasks, e.g., object detection~\cite{sr+detection} and face recognition~\cite{sr+face}.
However, image degradation is inevitable during the imaging process, and would consequently harm the following high-level tasks.
Image blur is a common image degradation in real-world scenarios, e.g., camera or object motion blur during the exposure time.
Furthermore, the blur process becomes more complicated due to the depth variations and occlusions around motion boundaries.
Therefore, real-world images often undergo the effects of non-uniform motion blur. 
The problems of SR and deblurring are often dealt separately, as each of them is known to be ill-posed. 
In this work, we aim at addressing the joint problem of generating clear HR images from given blurry LR inputs.
\figref{1} shows one blurry LR image that is affected by non-uniform blur. 
As existing SR algorithms~\cite{EDSR,srresnet,lapsrn,vdsr} cannot reduce motion blur well, the resulting HR image is blurry (see~\figref{1_sr_input} and~\ref{fig:1_sr}).
On the other hand, the state-of-the-art non-uniform deblurring methods~\cite{non_uniform_deblur1,db&optical,deepdeblur, deblurgan} generate clear images but cannot restore fine details and enlarge the spatial resolution (see~\figref{1_db_input} and~\ref{fig:1_db}).

With the advances of deep CNNs, state-of-the-art image super-resolution~\cite{EDSR,srresnet,lapsrn} and deblurring~\cite{deepdeblur,deblurgan} methods are built on end-to-end deep networks and achieve promising performance.
To jointly solve the image deblurring and super-resolution problems, a straightforward approach is to solve the two problems sequentially, i.e., performing image deblurring followed by super-resolution, or vice versa.
However, there are numerous issues with such approaches. 
First, a simple concatenation of two models is a sub-optimal solution due to error accumulation, i.e., the estimated error of the first model will be propagated and magnified in the second model.
Second, the two-step network does not fully exploit the correlation between the two tasks. 
For example, the feature extraction and image reconstruction steps are performed twice and result in computational redundancy.
As both the training and inference processes are memory and time-consuming, these approaches cannot be applied to resource-constrained applications. 
Several recent methods~\cite{scgan,facesr+attr,icassp18} jointly solve the image deblurring and super-resolution problems using end-to-end deep networks.
However, these methods either focus on domain-specific applications, e.g., face and text~\cite{scgan,facesr+attr} images, or address uniform Gaussian blur~\cite{icassp18}.
In contrast, we aim to handle natural images degraded by non-uniform motion blur, which is more challenging.

In this work, we propose a Gated Fusion Network (GFN) based on a dual-branch architecture to tackle this complex joint problem.
The proposed network consists of two branches: one branch aims to deblur the LR input image, and the other branch generates a clear HR output image.
In addition, we learn a gate module to adaptively fuse the features from the deblurring and super-resolution branches.
We fuse the two branches on the feature level rather than the intensity level to avoid the error accumulation from imperfect deblurred images.
The fused features are then fed into an image reconstruction module to generate clear HR output images.
Extensive evaluations demonstrate that the proposed model performs favorably against the combination of the state-of-the-art image deblurring and super-resolution methods as well as existing joint models.

The contributions of this work are summarized as follows:
\begin{compactitem}
\item To the best of our knowledge, this is the first CNN-based method to jointly solve non-uniform motion deblurring and generic single-image super-resolution.
\item We decouple the joint problem into two sub-tasks for better regularization.
We propose a dual-branch network to extract features for deblurring and super-resolution and learn a gate module to adaptively fuse the features for image reconstruction.
\item The proposed model is efficient and entails low computational cost as most operations are executed in the LR space.
Our model is much faster than the combination of the state-of-the-art SR and image deblurring models while achieving significant performance gain.
\end{compactitem}

\section{Related Work}
\label{sec:realted}
We aim to solve the image super-resolution and non-uniform motion deblurring problems jointly in one algorithm.
Both motion deblurring and image super-resolution are fundamental problems in computer vision.
In this section, we focus our discussion on recent methods for image super-resolution and motion deblurring closest to this work. 

{\flushleft \bf Image super-resolution.}
Single image super-resolution (SISR) is an ill-posed problem as there are multiple HR images correspond to the same LR input image.
Conventional approaches learn the LR-to-HR mappings using sparse dictionary~\cite{A+}, random forest~\cite{RFL} or self-similarity~\cite{SelfExSR}.
In recent years, CNN-based methods~\cite{srcnn,vdsr} have demonstrated significant performance improvement against conventional SR approaches.
Several techniques have been adopted, including the recursive learning~\cite{drrn,drcn}, pixel shuffling~\cite{pixelshuffle,EDSR}, and Laplacian pyramid network~\cite{lapsrn}.
In addition, some approaches use the adversarial loss~\cite{srresnet}, perceptual loss~\cite{Johnson-ECCV-2016} and texture loss~\cite{enhancenet} to generate more photo-realistic SR images.
As most SR algorithms assume that the LR images are generated by a simple downsampling kernel, e.g., bicubic kernel, they do not perform well when the input images suffer from unexpected motion blur.
In contrast, the proposed model performs super-resolution on LR blurry images.

{\flushleft \bf Motion deblurring.}
Conventional image deblurring approaches~\cite{cho2009fast,xu2013unnatural,pan2016blind,uniform_db1,uniform_db2,uniform_db3} assume that the blur is uniform and spatially invariant across the entire image.
However, due to the depth variation and object motion, real-world images typically contain non-uniform blur.
Several approaches solve the non-uniform deblurring problem by jointly estimating blur kernels with scene depth~\cite{kernel_estimate_non-uniform1,hu2014joint} or segmentation~\cite{db_related_1}.
As the kernel estimation step is computationally expensive, recent methods~\cite{text_db_cnn,non_uniform_deblur1,deepdeblur,non-uniform_kernel-free} learn deep CNNs to bypass the kernel estimation and efficiently solve the non-uniform deblurring problem.
Orest et al.~\cite{deblurgan} adopt the Wasserstein generative adversarial network (GAN) to generate realistic deblurred images and facilitate the object detection task.
When applying to LR blurry images, the above approaches cannot produce sufficient details and upscale the spatial resolution.

{\flushleft \bf Joint image super-resolution and motion deblurring.}
The joint problem is more challenging than the individual problem as the input images are extremely blurry and of low-resolution.
Some approaches~\cite{videoSRDB1,videoSRDB2,videoSRDB3} reconstruct sharp and HR images from a LR and blurry video sequence.
However, these methods rely on the optical flow estimation, which cannot be applied to the case where the input is a single image.
Xu et al.~\cite{scgan} train a generative adversarial network to super-resolve blurry face and text images.
As face and text images have relative small visual entropy, i.e., high patch recurrence within an image~\cite{zssr}, a small model can restore these category-specific images well.
Our experimental results show that the model of Xu et al.~\cite{scgan} does not have sufficient capacity to handle natural images with non-uniform blur.
Zhang et al.~\cite{icassp18} propose a deep encoder-decoder network (ED-DSRN) for joint image deblurring and super-resolution.
However, they focus on LR images degraded by uniform Gaussian blur.
In contrast, our model can handle complex non-uniform blur and use much fewer parameters to effectively restore clear HR images.

\section{Gated Fusion Network}
\label{sec:method}
In this section, we describe the architecture design, training loss functions, and implementation details of the proposed GFN for joint image deblurring and super-resolution.

\begin{figure*}[t]
\centering
\includegraphics[width=0.99\linewidth]{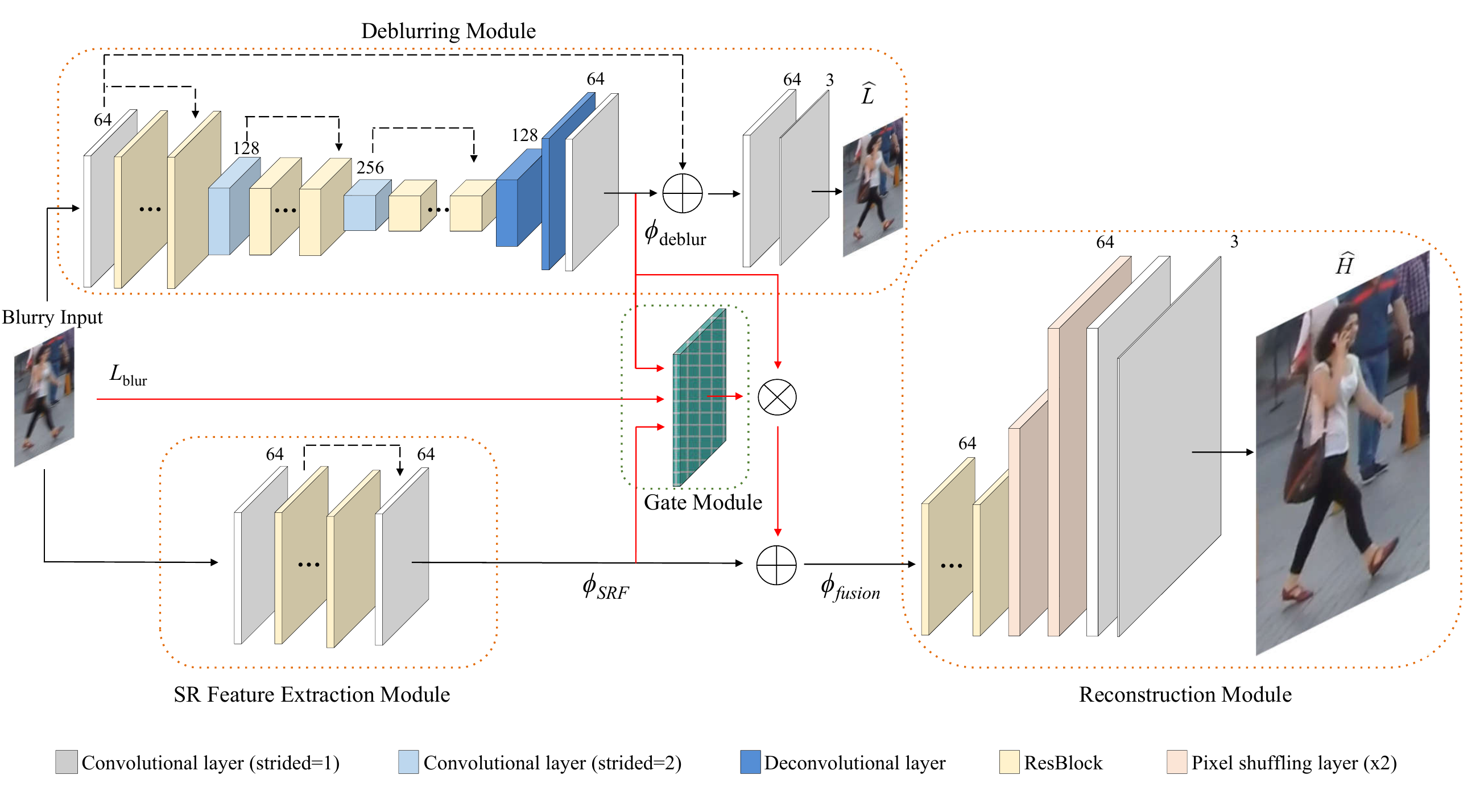}
\caption{\textbf{Architecture of the proposed GFN.}
Our model consists of four major modules:
deblurring module $G_{deblur}$, SR feature extraction module $G_{SRF}$, gate module $G_{gate}$, and reconstruction module $G_{recon}$.
The features extracted by $G_{deblur}$ and $G_{SRF}$ are fused by $G_{gate}$ and then fed into $G_{recon}$ to reconstruct the HR output image.
}
\label{fig:2}
\end{figure*}

\subsection{Network architecture}
Given a blurry LR image $L_{blur}$ as the input, our goal is to recover a sharp HR image $\widehat{H}$.
In this work, we consider the case of $4\times$ SR, i.e., the spatial resolution of $L_{blur}$ is $4\times$ smaller than that of $\widehat{H}$.
The proposed model is based on the dual-branch architecture~\cite{mask_rcnn,bi-net1,gated-net,li2016deep,zhang2017joint} and consists of four major modules:
(i) a deblurring module $G_{deblur}$ for extracting deblurring features and predicting a sharp LR image $\widehat{L}$, 
(ii) an SR feature extraction module $G_{SRF}$ to extract features for image super-resolution, 
(iii) a gate module $G_{gate}$ to estimate a weight map for blending the deblurring and super-resolution features, 
and (iv) a reconstruction module $G_{recon}$ to reconstruct the final HR output image.
An overview of the proposed model is illustrated in~\figref{2}.

\paragraph{Debluring module.}
The deblurring module aims to restore a sharp LR image $\widehat{L}$ from the input blurry LR image $L_{blur}$.
Different from the encoder-decoder architecture in ~\cite{deepvideo}, we adopt an asymmetric residual encoder-decoder architecture in our deblurring module to enlarge the receptive field.
The encoder consists of three scales, where each scale has six ResBlocks~\cite{EDSR} followed by a strided convolutional layer to downsample the feature maps by $1/2\times$.
The decoder has two deconvolutional layers to enlarge the spatial resolution of feature maps.
Finally, we use two additional convolutional layers to reconstruct a sharp LR image $\widehat{L}$.
We denote the output features of the decoder by $\phi_{deblur}$, which are later fed into the gate module for feature fusion.

\paragraph{Super-resolution feature extraction module.}
We use eight ResBlocks~\cite{EDSR} to extract high-dimensional features for image super-resolution.
To maintain the spatial information, we do not use any pooling or strided convolutional layers.
We denote the extracted features as $\phi_{SRF}$.

\begin{figure}[tb]
\centering
\begin{tabular}{cccc}
  \includegraphics[width=0.25\linewidth]{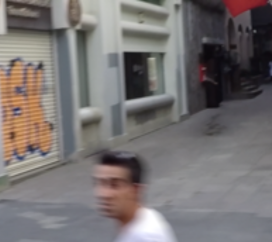} & \hspace{-4mm}
  \includegraphics[width=0.25\linewidth]{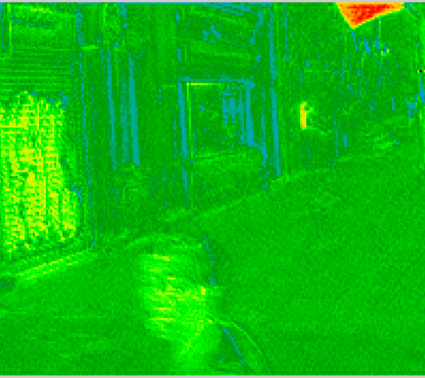} & \hspace{-4mm}
  \includegraphics[width=0.25\linewidth]{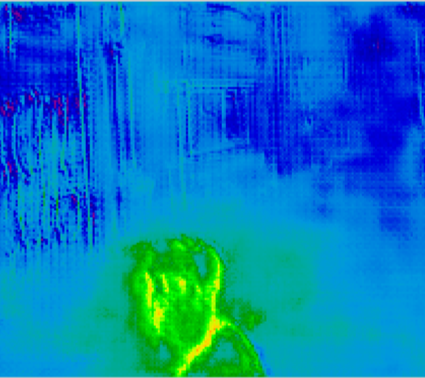} &
\hspace{-5mm}
  \includegraphics[width=0.071\linewidth]{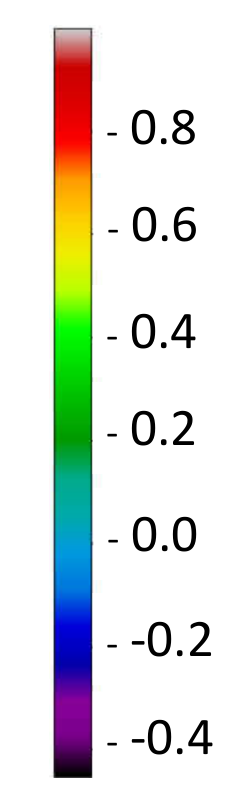}\\
  (a) Input image & (b) $\phi_{SRF}$ & (c) $\phi_{deblur}$
\end{tabular}
\caption{
\textbf{Feature response of the super-resolution features $\phi_{SRF}$ and the deblurring features $\phi_{deblur}$.}
%
While the super-resolution features contain more spatial details, the deblurring features have higher response on large motion.
%
}
\label{fig:Figure3}
\end{figure}

\paragraph{Gate module.}
In~\figref{Figure3}, we visualize the responses of $\phi_{deblur}$ and $\phi_{SRF}$.
While the SR features contain spatial details of the input image (as shown on the wall of~\figref{Figure3}(b)), the deblurring features have high response on regions of large motion (as shown on the moving person of~\figref{Figure3}(c)).
Thus, the responses of $\phi_{deblur}$ and $\phi_{SRF}$ complement each other, especially on blurry regions.
Inspired by the fact that gate structures can be used to discover feature importance for multimodal fusion~\cite{lstm,GFN_dehaze}, we propose to learn a gate module to adaptively fuse the two features.
Unlike~\cite{GFN_dehaze} that predicts the weight maps of three derived inputs, our gate module enables local and contextual feature fusion from two independent branches, and thus it is expected to selectively merge sharp features $\phi_{fusion}$ from $\phi_{deblur}$ and $\phi_{SRF}$ in this work.
Our gate module, $G_{gate}$, takes as input the set of $\phi_{deblur}$, $\phi_{SRF}$, and the LR blurry image $L_{blur}$, and generate a pixel-wise weight maps for blending $\phi_{deblur}$ and $\phi_{SRF}$:
\begin{equation}\label{eqn2}
	\phi_{fusion} = G_{gate}(\phi_{SRF}, \phi_{deblur}, L_{blur})  \otimes \phi_{deblur} + \phi_{SRF}, 
\end{equation}
where $\otimes$ denotes the element-wise multiplication.
The gate module consists of two convolutional layers with the filter size of $3 \times 3$ and $1 \times 1$, respectively.

\paragraph{Reconstruction module.}
The fused features $\phi_{fusion}$ from the gate module are fed into eight ResBlocks and two pixel-shuffling layers~\cite{pixelshuffle} to enlarge the spatial resolution by $4\times$.
We then use two final convolutional layers to reconstruct an HR output image $\widehat{H}$.
We note that most of the network operations are performed in the LR feature space.
Therefore, the proposed model entails low computational cost in both training and inference phases.

\subsection{Loss functions}
The proposed network generates two output images: a deblurred LR image $\widehat{L}$ and a sharp HR image $\widehat{H}$.
In our training data, each blurry LR image $L_{blur}$ has a corresponding ground truth HR image $H$ and a ground truth LR image $L$, which is the bicubic downsampled image of $H$.
Therefore, we train our network by jointly optimizing a super-resolution loss and a deblurring loss:
\begin{equation}\label{eqn3}
	\min~ \mathcal{L}_{SR}(\widehat{H}, H) + \alpha \mathcal{L}_{deblur}(\widehat{L}, L),
\end{equation}
where $\alpha$ is a weight to balance the two loss terms.
We use the pixel-wise MSE loss function for both $\mathcal{L}_{SR}$ and $\mathcal{L}_{deblur}$, and empirically set $\alpha = 0.5$.

\subsection{Implementation and training details}
In the proposed network, the filter size is set as $7 \times 7$ in the first and the last convolutional layers, $4 \times 4$ in the deconvolutional layers, and $3 \times 3$ in all other convolutional layers.
We initialize all learnable parameters from a Gaussian distribution.
We use the leaky rectified linear unit (LReLU) with a negative slope of 0.2 as the activation function.
As suggested in~\cite{EDSR}, we do not use any batch normalization layers in order to retain the range flexibility of features.
More implementation details of the network can be found in the supplementary due to space limitation.

To facilitate the network training, we adopt a two-step training strategy.
First, we disable the gate module and fuse the deblurring feature and SR feature directly.
%
We train the deblurring/super-resolution/reconstruction modules from scratch for 60 epochs.
We initialize the learning rate to 1e-4 and multiply by 0.1 for every 30 epochs.
Second, we include the gate module and train the entire network for another 50 epochs.
The learning rate is set to 5e-5 and multiplied by 0.1 for every 25 epochs.
We use a batch size of 16 and augment the training data by random rotation and horizontal flipping.
We use the ADAM solver~\cite{adam} with $\beta_1 = 0.9$ and $\beta_2 = 0.999$ to optimize our network.
The training process takes about 4 days on an NVIDIA 1080Ti GPU card.
The source code can be found at \href{http://xinyizhang.tech/bmvc2018}{http://xinyizhang.tech/bmvc2018} .

\section{Experimental Results}
\label{sec:experiment}

In this section, we first describe the dataset for training our network, present quantitative and qualitative comparisons with state-of-the-art approaches, and finally analyze and discuss several design choices of the proposed model.

\subsection{Training dataset}
We use the GOPRO dataset~\cite{deepdeblur} to generate the training data for the joint SR and deblurring problem.
The GOPRO dataset contains 2103 blurry and sharp HR image pairs for training.
To generate more training data, we resize each HR image pair with three random scales within $\left[0.5, 1.0\right]$.
We then crop HR image pairs into $256 \times 256$ patches with a stride of 128 to obtain blurry HR patches $H_{blur}$ and sharp HR patches $H$.
Finally, we use the bicubic downsampling to downscale $H_{blur}$ and $H$ by $4\times$ to generate blurry LR patches $L_{blur}$ and sharp LR patches ${L}$, respectively.
In total, we obtain 10,7584 triplets of $\left\{L_{blur}, L, H\right\}$ for training (the blurry HR patches $H_{blur}$ are discarded during training).
We refer the generated dataset as LR-GOPRO in the following content.

\subsection{Comparisons with state-of-the-arts}
We compare the proposed GFN with several algorithms, including the state-of-the-art SR methods~\cite{srresnet,EDSR}, the joint image deblurring and SR approaches~\cite{scgan,icassp18}, and the combinations of SR algorithms~\cite{srresnet,EDSR} and non-uniform deblurring algorithms~\cite{deepdeblur,deblurgan}.
For fair comparisons, we re-train the models of SCGAN~\cite{scgan}, SRResNet~\cite{srresnet}, and ED-DSRN~\cite{icassp18} on our training dataset.

\begin{figure}[tb]
\centering
\subfigure[PSNR vs. inference time]{\label{fig:psnr_time}
\begin{minipage}[t]{0.48\linewidth}
  \centering
  \includegraphics[height=5cm]{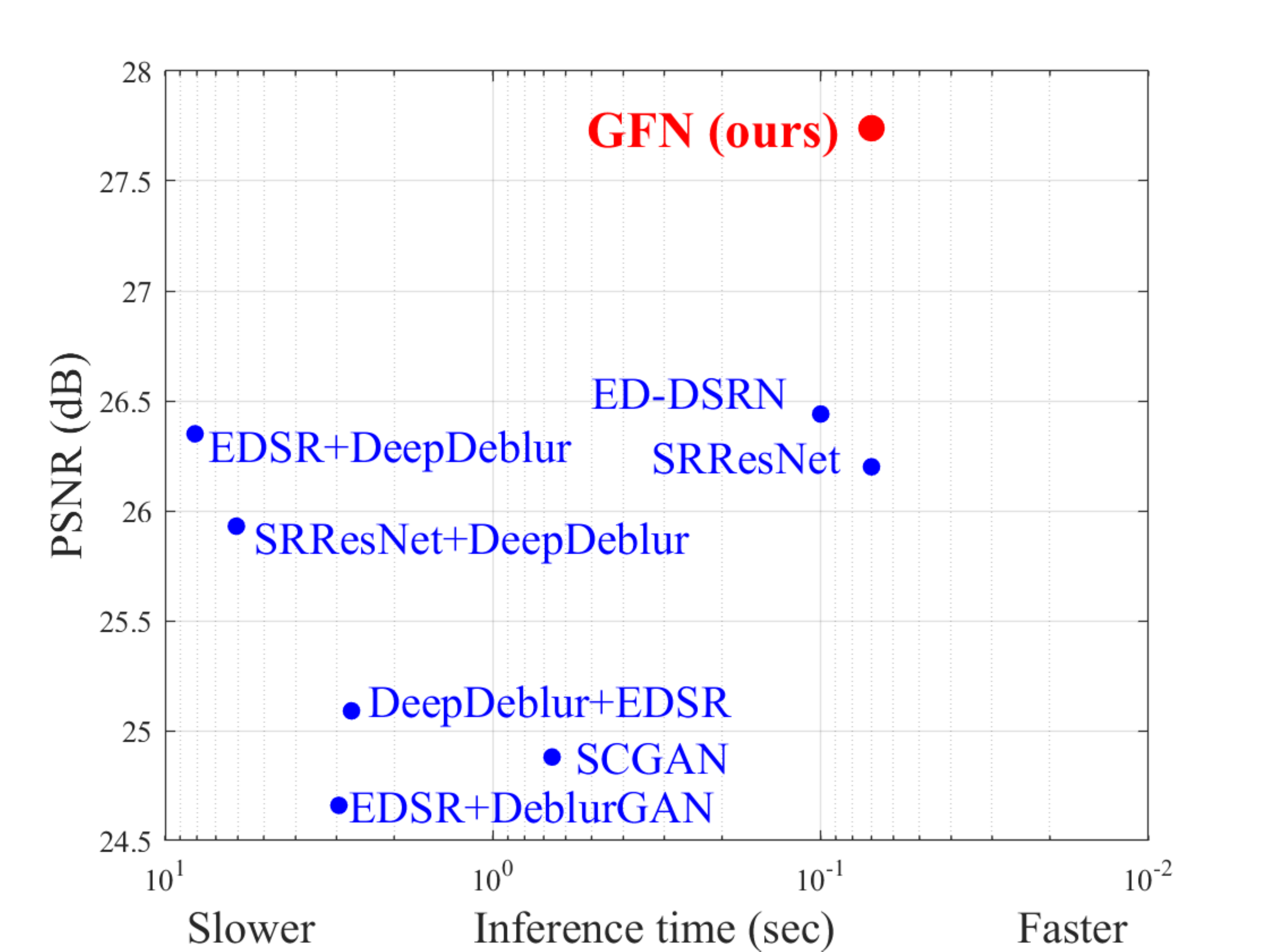}
  \end{minipage}
}
\subfigure[PSNR vs. number of parameters]{\label{fig:psnr_params}
\begin{minipage}[t]{0.48\linewidth}
  \centering
  \includegraphics[height=5cm]{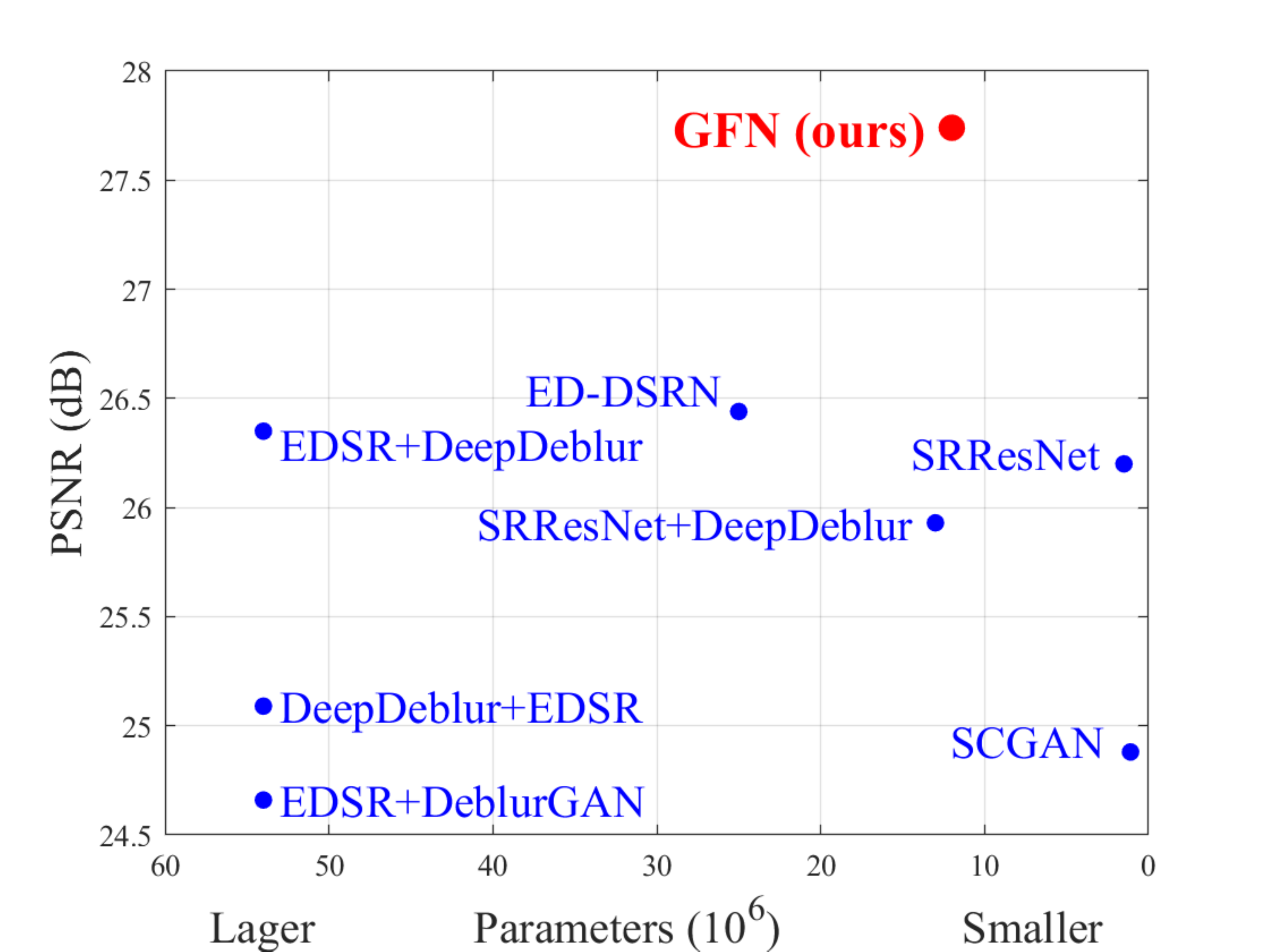}
  \end{minipage}
}
\caption{
\textbf{Performance versus inference time and model parameters.}
The results are evaluated on the LR-GOPRO dataset.
}
\label{fig:psnr_time_params}
\end{figure}

We use the bicubic downsampling to generate blurry LR images from the test set of the GOPRO dataset~\cite{deepdeblur} and the K\"ohler dataset~\cite{kohler} for evaluation.
\tabref{table_results} shows the quantitative evaluation in terms of PSNR, SSIM, and average inference time.
The tradeoff between image quality and efficiency is better visualized in~\figref{psnr_time_params}.
The proposed GFN performs favorably against existing methods on both datasets and maintains a low computational cost and execution time.
While the re-trained SCGAN and SRResNet perform better than their pre-trained models, both methods cannot handle the complex non-uniform blur well due to their small model capacity.
The ED-DSRN method uses more network parameters to achieve decent performance.
However, the single-branch architecture of ED-DSRN is less effective than the proposed dual-branch network.

\begin{table*}[t]
\small
\centering
\caption{
\textbf{Quantitative comparison with state-of-the-art methods.}
The methods with a $\star$ sign are trained on our LR-GOPRO training set.
{\color{red}Red texts} indicate the best performance.
The proposed GFN performs favorably against existing methods while maintaining a small model size and fast inference speed.
}
\label{tab:table_results}
\begin{tabular}{rcll}
\hline
&\multicolumn{1}{l}{} & \multicolumn{1}{c}{LR-GOPRO $4\times$}   & \multicolumn{1}{c}{LR-K\"ohler $4\times$} 
\\
\multirow{-2}{*}{Method}    &\multirow{-2}{*}{\#Params}     & \multicolumn{1}{c}{PSNR~/~SSIM~/~Time~(s)}    & \multicolumn{1}{c}{PSNR~/~SSIM~/~Time~(s)}      
\\ 
\hline
SCGAN~\cite{scgan}        &1.1M      &22.74~/~0.783~/~0.66  &23.19~/~0.763~/~0.45
\\
SRResNet~\cite{srresnet}  &1.5M    &24.40~/~0.827~/~{\color{red}0.07} &24.81~/~0.781~/~{\color{red}0.05}
\\
EDSR~\cite{EDSR}          &43M     &24.52~/~0.836~/~2.10    &24.86~/~0.782~/~1.43
\\
SCGAN${}^{\star}$~\cite{scgan}	&1.1M &24.88~/~0.836~/~0.66 &24.82~/~0.795~/~0.45
\\  
SRResNet${}^{\star}$~\cite{srresnet} &1.5M	&26.20~/~0.818~/~{\color{red}0.07} &25.36~/~0.803~/~{\color{red}0.05}
\\  
ED-DSRN${}^{\star}$~\cite{icassp18}	&25M &26.44~/~0.873~/~0.10 &25.17~/~0.799~/~0.08
\\
\hline
DB~\cite{deepdeblur} + SR~\cite{srresnet} &13M &24.99~/~0.827~/~0.66 &25.12~/~0.800~/~0.55
\\
SR~\cite{srresnet} + DB~\cite{deepdeblur} &13M  &25.93~/~0.850~/~6.06 &25.15~/~0.792~/~4.18
\\
DB~\cite{deblurgan} + SR~\cite{srresnet}  &13M &21.71~/~0.686~/~0.14 &21.10~/~0.628~/~0.12
\\
SR~\cite{srresnet} + DB~\cite{deblurgan}  &13M &24.44~/~0.807~/~0.91  &24.92~/~0.778~/~0.54
\\ 
DB~\cite{deblurgan} + SR~\cite{EDSR}      &54M &21.53~/~0.682~/~2.18 &20.74~/~0.625~/~1.57
\\
SR~\cite{EDSR} + DB~\cite{deblurgan}  &54M &24.66~/~0.827~/~2.95 &25.00~/~0.784~/~1.92
\\ 
DB~\cite{deepdeblur} + SR~\cite{EDSR}     &54M  &25.09~/~0.834~/~2.70 &25.16~/~0.801~/~2.04
\\
SR~\cite{EDSR} + DB~\cite{deepdeblur}  &54M &26.35~/~0.869~/~8.10    &25.24~/~0.795~/~5.81
\\ 
\hline
GFN~(ours) &12M & {\color{red} 27.74}~/~{\color{red} 0.896}~/~{\color{red}0.07}    & {\color{red} 25.72}~/~{\color{red} 0.813}~/~{\color{red}0.05}
\\
\hline
\end{tabular}
\end{table*}

The concatenations of SR~\cite{srresnet,EDSR} and deblurring methods~\cite{deepdeblur,deblurgan} are typically less effective due to the error accumulation.
We note that the concatenations using the SR-first strategy, i.e., performing SR followed by image deblurring, typically have better performance than that using the DB-first strategy, i.e., performing image deblurring followed by SR.
However, the SR-first strategy entails heavy computational cost as the image deblurring step is performed in the HR image space.
Compared with the best-performing combination of EDSR~\cite{EDSR} and DeepDeblur~\cite{deepdeblur} methods, the proposed GFN is $116\times$ faster and has $78\%$ fewer model parameters.

\begin{figure}[tb]
\small
\centering
\begin{tabular}{cccc}
	\hspace{-3mm}
 	\includegraphics[width=0.245\linewidth]{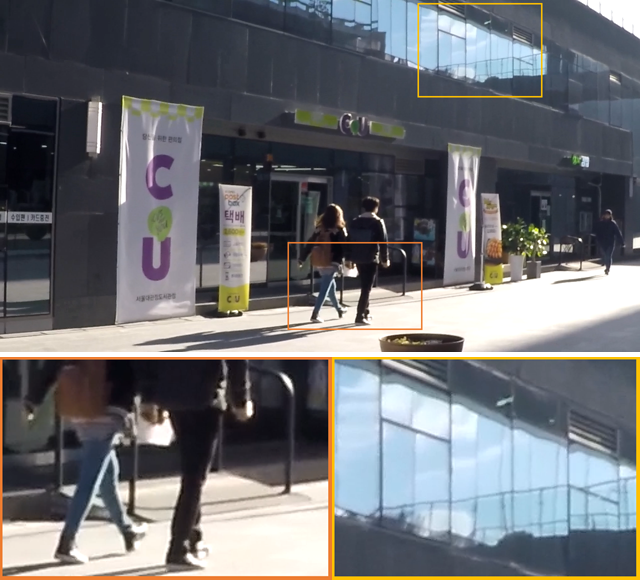} & \hspace{-4mm}
  	\includegraphics[width=0.245\linewidth]{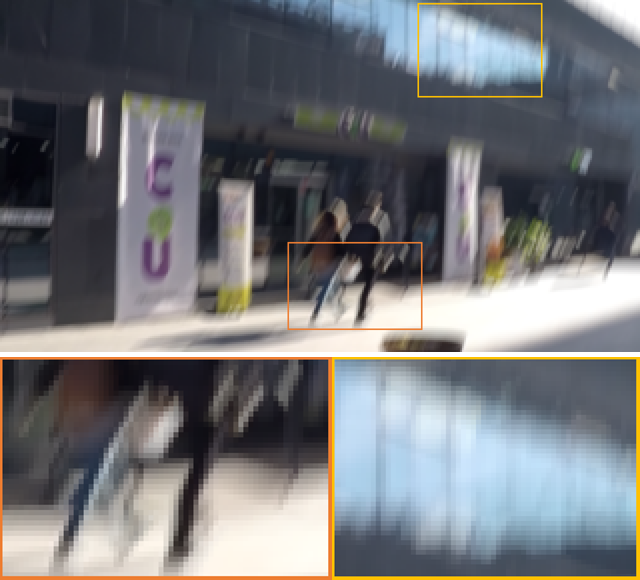} & \hspace{-4mm}
  	\includegraphics[width=0.245\linewidth]{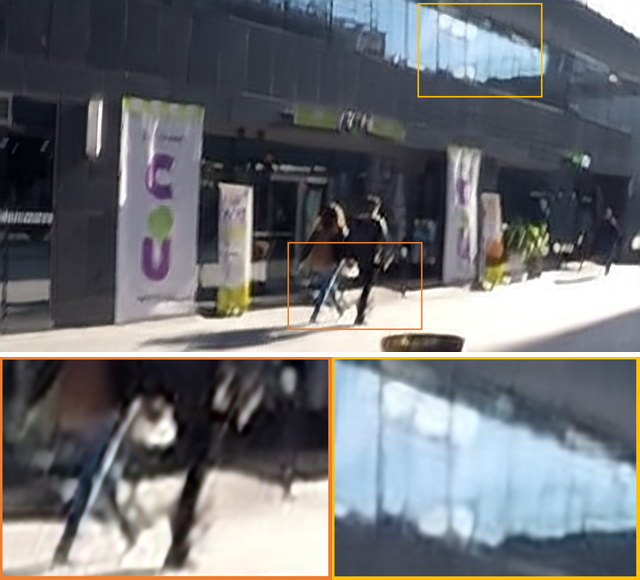} & \hspace{-4mm}
  	\includegraphics[width=0.245\linewidth]{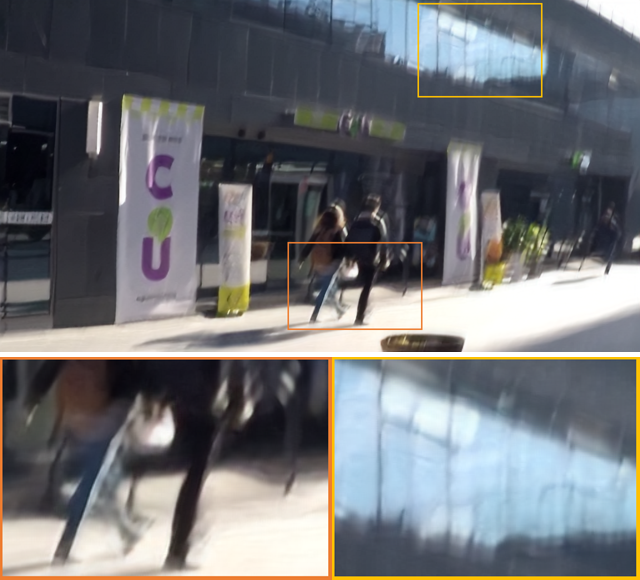}
	\vspace{-0.5mm}
	\\
	\hspace{-3mm} 
    (a) Ground-truth HR & \hspace{-4mm}
	(b) Blurry LR input & \hspace{-4mm}
    (c) DB~\cite{deepdeblur} + SR~\cite{EDSR} & \hspace{-4mm}
    (d) SR~\cite{EDSR} + DB~\cite{deepdeblur}
	\\
	\hspace{-3mm} 
    PSNR / SSIM & \hspace{-4mm}
    21.04 / 0.787 & \hspace{-4mm}
    24.58 / 0.846 & \hspace{-4mm}
    25.04 / 0.876
	\\
	\hspace{-3mm}
  	\includegraphics[width=0.245\linewidth]{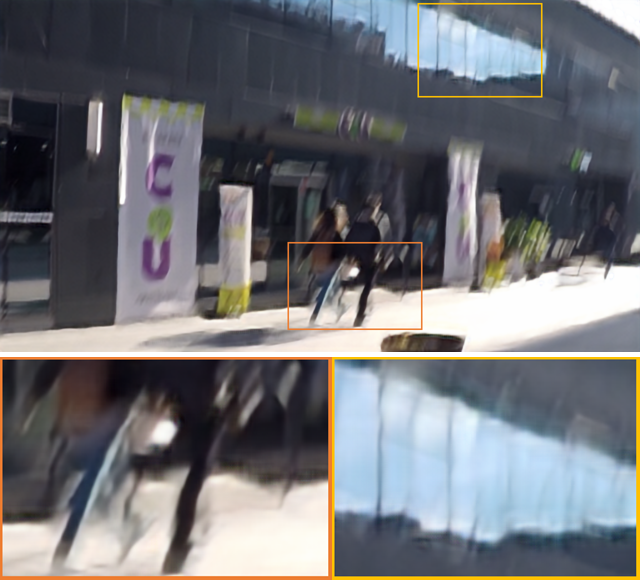} & \hspace{-4mm}
  	\includegraphics[width=0.245\linewidth]{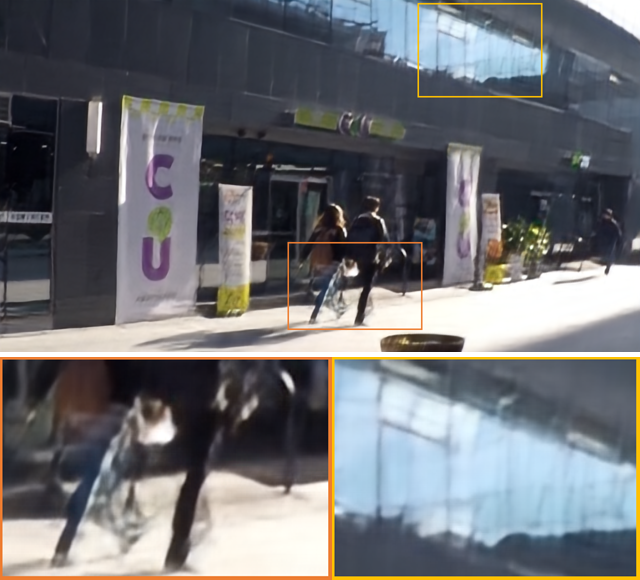} & \hspace{-4mm}
  	\includegraphics[width=0.245\linewidth]{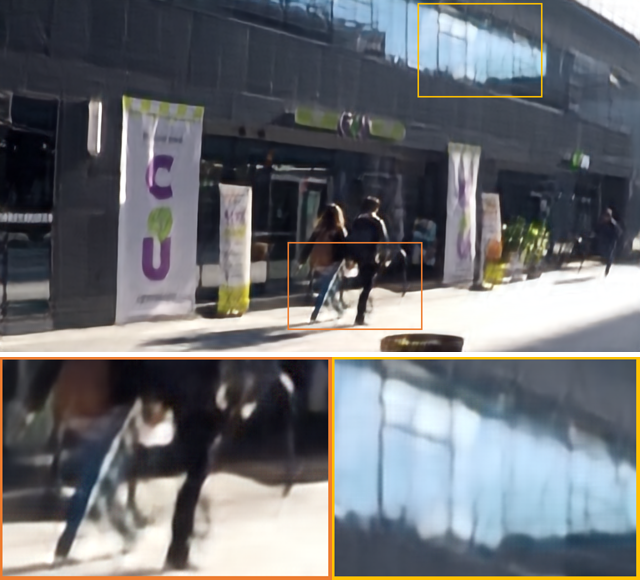} & \hspace{-4mm}
  	\includegraphics[width=0.245\linewidth]{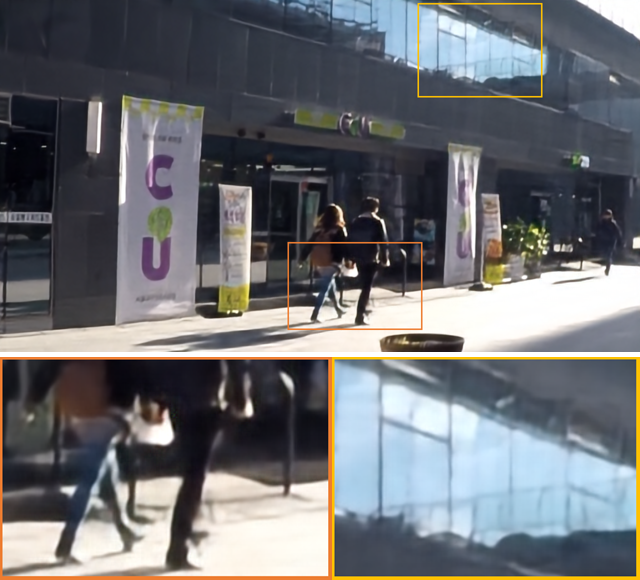}
  	\vspace{-0.5mm}
	\\
	\hspace{-3mm} 
    (e) SCGAN~\cite{scgan} & \hspace{-4mm}
    (f) ED-DSRN~\cite{icassp18} & \hspace{-4mm} 
    (g) SRResNet\cite{srresnet} & \hspace{-4mm}
    (h) GFN (ours)
	\\
	\hspace{-3mm} 
    23.00 / 0.835 & \hspace{-4mm}
    26.07 / 0.896 & \hspace{-4mm}
    25.63 / 0.881 & \hspace{-4mm}
    29.02 / 0.929
\end{tabular}
\caption{
\textbf{Visual comparison on the LR-GOPRO dataset.}
The proposed method generates sharp HR images with more details.
}
\label{fig:visual_results_GOPRO}
\end{figure}

\begin{figure}[tb]
\small
\centering
\vspace{-1mm}
\begin{tabular}{cccc}
\hspace{-3mm}
  \includegraphics[width=0.245\linewidth]{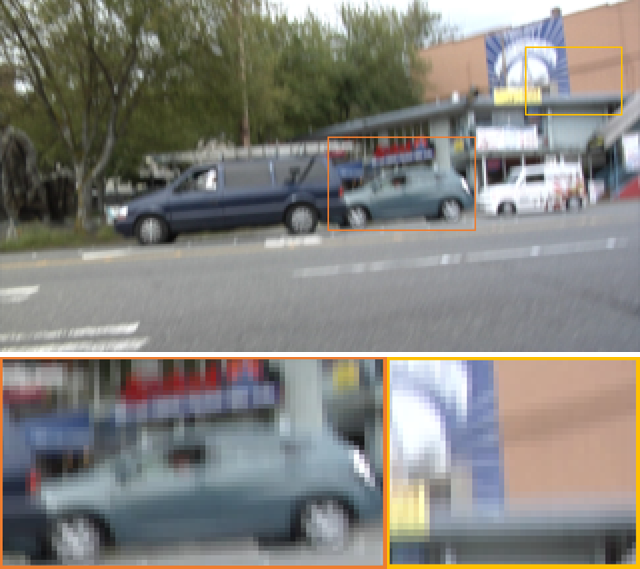} & \hspace{-4mm}
  \includegraphics[width=0.245\linewidth]{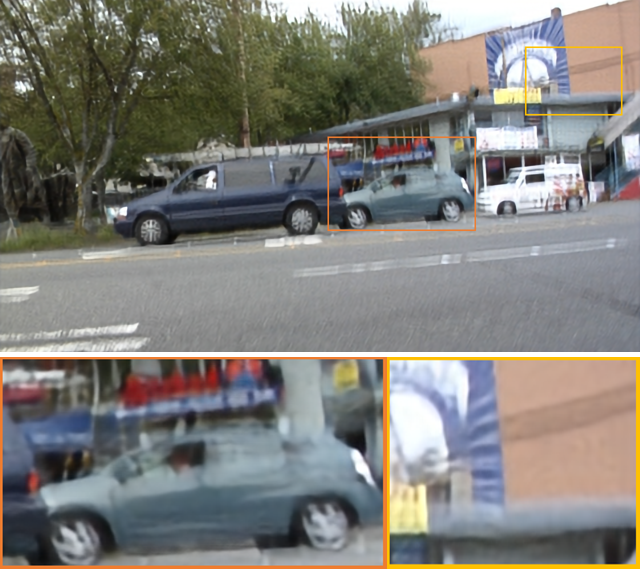} & \hspace{-4mm}
  \includegraphics[width=0.245\linewidth]{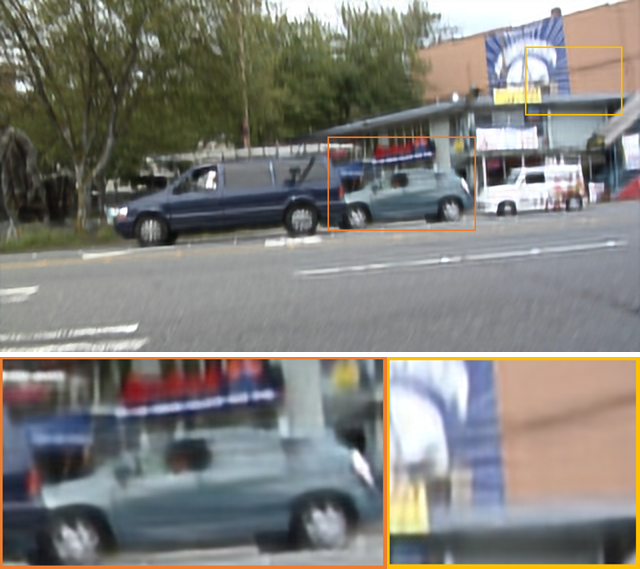} &
\hspace{-4mm}
  \includegraphics[width=0.245\linewidth]{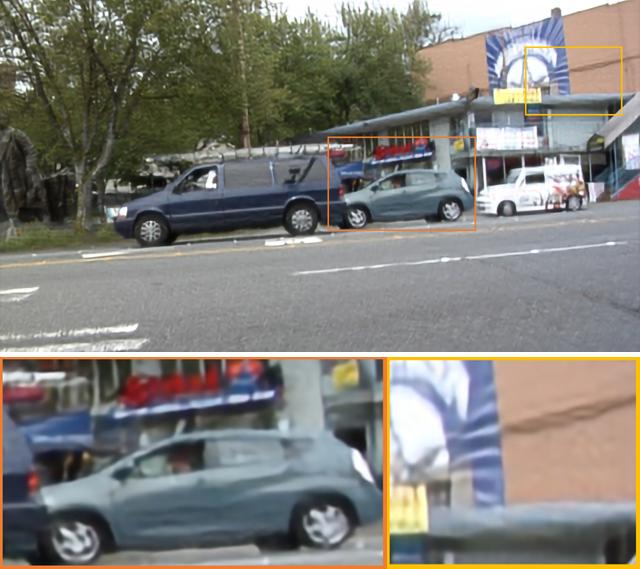}
\vspace{-0.5mm}
\\
\hspace{-4mm} (a) Blurry LR input & (b) ED-DSRN~\cite{icassp18} & (c) SRResNet\cite{srresnet} &(d) GFN (ours) 
\end{tabular}
\caption{
\textbf{Visual comparison on the real blurry image dataset~\cite{deepvideo}.}
Our GFN is more robust to outliers in real images and generates a sharper result than state-of-the-art methods~\cite{icassp18,srresnet}.
}
\label{fig:visual_results_real}
\end{figure}

We present the qualitative results on the LR-GOPRO dataset in~\figref{visual_results_GOPRO} and the results on a real blurry image in~\figref{visual_results_real}.
The methods using the concatenation scheme, e.g., DeepDeblur~\cite{deepdeblur} + EDSR~\cite{EDSR} and EDSR~\cite{EDSR} + DeepDeblur~\cite{deepdeblur}, often introduce undesired artifacts due to the error accumulation problem.
Existing joint SR and deblurring methods~\cite{icassp18,srresnet,scgan} cannot reduce the non-uniform blur well.
In contrast, the proposed method generates clear HR images with more details.

\begin{table*}[]
\small
\centering
\caption{
\textbf{Analysis on key components in the proposed GFN.}
All the models are trained on the LR-GOPRO dataset with the same hyper-parameters.
}
\label{table_Ablation}
\begin{tabular*}{1\textwidth}{@{\extracolsep{\fill}}r|p{1.2cm}<{\centering}|p{1.1cm}<{\centering} p{1.1cm}<{\centering} p{1.1cm}<{\centering} p{1.1cm}<{\centering} p{1.1cm}<{\centering}|p{0.7cm}<{\centering}}
\hline
\multirow{-1}{*}{Modifications}  &SRResNet &Model-1  &Model-2 &Model-3 &Model-4 &Model-5 &GFN   
\\ 
\hline
deblurring loss         &\checkmark &\checkmark	& &\checkmark	& &\checkmark	&\checkmark    
\\
deblurring module       &	&\checkmark	&\checkmark	&\checkmark	&\checkmark	&\checkmark	&\checkmark      
\\
dual-branch            &	&	&	&\checkmark	&\checkmark	&\checkmark	&\checkmark            
\\
feature level    &	&	&	&	&\checkmark	&\checkmark	&\checkmark     
\\
gate module             &	&	&	&	&	&	&\checkmark         
\\
SR-first                &	&	&\checkmark 	&	&	&	&  \\
\hline
PSNR                    &26.20	&26.82	&27.00	&27.01	&27.39	&27.53	&27.74	
\\
Time (s)                &0.05	&0.09	&0.57	&0.10	&0.05	&0.05	&0.05
\\
\hline
\end{tabular*}
\label{tab:ablation}
\vspace{-2mm}
\end{table*}

\subsection{Analysis and discussion}
\label{subsec:Ablation}
The proposed GFN consists of several key components:
(1) using a \textnormal{deblurring loss} for regularization; 
(2) using a \textnormal{deblurring module} to extract the deblurring features; 
(3) using a \textnormal{dual-branch} architecture instead of a concatenation of SR and deblurring models; 
(4) fusing features in the \textnormal{feature level} instead of the intensity level; 
(5) using a \textnormal{gate module} to adaptive fuse SR and deblurring features.
The training schemes of these models are presented in the supplementary material.
Here we discuss the performance contribution of these components.

We first train a baseline model by introducing a deblurring loss to the SRResNet.
We then include the deblurring module to the baseline model using two sequential strategies: DB-first (Model-1) and \textnormal{SR-first} (Model-2).
In~\tabref{ablation}, the deblurring module shows significant performance improvement over the baseline SRResNet model.
The SR-first model achieves better performance but has a slower execution speed.
However, the concatenation of SR and deblurring modules might be sub-optimal due to the error accumulation.
Next, we employ the dual-branch architecture and fuse the features in the intensity level (Model-3) and the feature level (Model-5).
Compared with Model-3, Model-5 achieves 0.5 dB performance improvement and runs faster.
We also train a model without the deblurring loss (Model-4).
Experimental results show that the deblurring loss helps the network converge faster and obtain improved performance (0.14 dB).
Finally, we show that the learned gate module can further boost the performance by 0.2 dB.

\section{Conclusions}
In this paper, we propose an efficient end-to-end network to recover sharp high-resolution images from blurry low-resolution input.
The proposed network consists of two branches to extract blurry and sharp features more effectively.
The extracted features are fused through a gate module, and are then used to reconstruct the final results.
The network design decouples the joint problem into two restoration tasks, and enables efficient training and inference.
Extensive evaluations show that the proposed algorithm performs favorably against the state-of-the-art methods in terms of visual quality and runtime.

\section*{Acknowledgement}
This work is partially supported by National Science and Technology Major Project (No. 2018ZX01008103), NSF CARRER (No.1149783), and gifts from Adobe and Nvidia.
\\
\\
\\
\\
\\
\\
\\
\\

\bibliography{egbib}
\end{document}